\pdfoutput=1

\documentclass[11pt,a4paper]{article}
\usepackage[hyperref]{acl2021}
\usepackage{times}
\usepackage{tikz}
\usepackage{subfig}
\usepackage{xcolor}
\usepackage{tcolorbox}
\usepackage{latexsym}

\usepackage{multirow}
\usepackage{booktabs}
\usepackage{pgfplots}
\usepackage{amssymb}
\usepackage{array}
\usepackage{xspace}
\usepackage{pgfplotstable}
\usepackage{pgf}

\usetikzlibrary{plotmarks}
\usetikzlibrary{backgrounds}
\usetikzlibrary{fit}
\usetikzlibrary{decorations}
\usetikzlibrary{decorations.markings}
\usetikzlibrary{positioning}

\definecolor{myblue}{RGB}{31,120,180}
\definecolor{mygreen}{RGB}{51,160,44}
\definecolor{myred}{RGB}{227,26,28}

\definecolor{iyellow}{RGB}{255,250,205}
\definecolor{ipurple}{RGB}{230,230,250}

\definecolor{upurple}{RGB}{155,89,182}
\definecolor{ublue}{RGB}{52,152,219}
\definecolor{ured}{RGB}{231,76,60}
\definecolor{udark}{RGB}{77,153,77}
\definecolor{ugreen}{RGB}{46,204,113}

\usepackage{microtype}

\aclfinalcopy 
\def\aclpaperid{1056} 

\title{ODE Transformer: An Ordinary Differential Equation-Inspired Model for Neural Machine Translation}

\author{
  Bei Li$^1$,
  Quan Du$^{1,2}$,
  Tao Zhou$^1$,
  Shuhan Zhou$^1$,
  Xin Zeng$^1$,
  \textbf{Tong Xiao$^{1,2}$\thanks{\xspace\xspace Corresponding author.}},
  \textbf{and Jingbo Zhu$^{1,2}$}\\
  $^{1}$NLP Lab, School of Computer Science and Engineering, \\
  Northeastern University, Shenyang, China\\
  $^{2}$NiuTrans Research, Shenyang, China \\
  {\tt
        \{libei\_neu,duquanneu,zhoutao\_neu,zsh\_neu,iszengxin\}@outlook.com 
  }\\
  {\tt
        \{xiaotong,zhujingbo\}@mail.neu.edu.cn
  }
}

\date{}

\begin{document}
\maketitle
\begin{abstract}
It has been found that residual networks are an Euler discretization of solutions to Ordinary Differential Equations (ODEs). In this paper, we explore a deeper relationship between Transformer and numerical methods of ODEs. We show that a residual block of layers in Transformer can be described as a higher-order solution to ODEs. This leads us to design a new architecture (call it ODE Transformer) analogous to the Runge-Kutta method that is well motivated in ODEs. As a natural extension to Transformer, ODE Transformer is easy to implement and parameter efficient. Our experiments on three WMT tasks demonstrate the genericity of this model, and large improvements in performance over several strong baselines. It achieves 30.76 and 44.11 BLEU scores on the WMT'14 En-De and En-Fr test data. This sets a new state-of-the-art on the WMT'14 En-Fr task.

\end{abstract}

\section{Introduction}

Residual networks have been used with a great success as a standard method of easing information flow in multi-layer neural models \cite{he2016deep,vaswani2017attention}. Given an input $y_t$, models of this kind define the output of a layer at depth $t$ to be:

\begin{eqnarray}
y_{t+1} & = & y_t + F(y_t, \theta_t) \label{eq:residual}
\end{eqnarray}

\noindent where $F(\cdot,\cdot)$ is the function of the layer and $\theta_t$ is its parameter. Interestingly, recent work in machine learning \cite{weinan2017proposal,yiping2018beyond,haber2018learning,chang2018reversible,ruthottohaber2019} points out that Eq. (\ref{eq:residual}) is an Euler discretization of the Ordinary Differential Equation (ODE), like this:

\begin{eqnarray}
\frac{\mathrm{d} y(t)}{\mathrm{d} t} & = & F(y(t), \theta(t)) \label{eq:ode}
\end{eqnarray}

\noindent where $y(t)$ and $\theta(t)$ are continuous with respect to $t$. In this way, we can call Eq. (\ref{eq:residual}) an \textit{ODE block}. This finding offers a new way of explaining residual networks in the view of numerical algorithms. Then, one can think of a multi-layer network as applying the Euler method (i.e., Eq. (\ref{eq:residual})) to solve Eq. (\ref{eq:ode}) subject to the initial conditions $y(0)=y_0$ and $\theta(0)=\theta_0$.

The solution of Eq. (\ref{eq:ode}) has a sufficiently low error bound (call it a \textit{stable solution}) only if $\theta(t)$ changes slow along $t$ \cite{haberruthotto2017,chen2018neural}. But this assumption does not always hold for state-of-the-art natural language processing (NLP) systems, in which models are non-linear and over-parameterized. For example, language modeling and machine translation systems learn quite different parameters for different layers, especially when the layers are close to the model input \cite{vaswani2017attention,dai-etal-2019-transformer}. Also, truncation errors are nonnegligible for the Euler method because it is a first-order approximation to the true solution \cite{he2019ode}. These problems make the situation worse, when more layers are stacked and errors are propagated through the neural network. It might explain why recent Machine Translation (MT) systems cannot benefit from extremely deep models \cite{wang-etal-2019-learning,liu-etal-2020-understanding,wei-etal-2020-multiscale,li-etal-2020-shallow}.

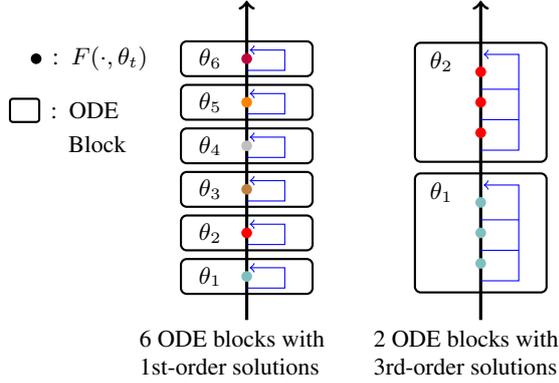
\begin{figure}[t!]
\centering
\begin{tikzpicture}

\newlength{\seg}
\setlength{\seg}{1.5em}
\newlength{\subseg}
\setlength{\subseg}{8em}
\newlength{\resseg}
\setlength{\resseg}{1.3em}

\tikzstyle{circlenode}=[circle,minimum size=4pt,inner sep=0,fill=blue!80];
\tikzstyle{odenode}=[draw,minimum height=1.2em,minimum width=4.5em,inner sep=0,thick,rounded corners=.2em];

\draw [->,very thick] (0,0) -- (0,11em);
\node [anchor=north] (cap01) at (-0.5em,0) {\footnotesize{6 ODE blocks with}};
\node [anchor=north west] (cap02) at ([yshift=0.3em]cap01.south west) {\footnotesize{1st-order solutions}};

\node [circlenode,fill=teal!50] (node01) at (0,0+1*\seg) {};
\node [circlenode,fill=red] (node02) at (0,0+2*\seg) {};
\node [circlenode,fill=brown] (node03) at (0,0+3*\seg) {};
\node [circlenode,fill=gray!50] (node04) at (0,0+4*\seg) {};
\node [circlenode,fill=orange] (node05) at (0,0+5*\seg) {};
\node [circlenode,fill=purple] (node06) at (0,0+6*\seg) {};

\foreach \i in {1,...,6}
{
    \node [odenode] (layer0\i) at (node0\i) {};
    \node [anchor=west] (label0\i) at ([xshift=0.3em]layer0\i.west) {\footnotesize{$\theta_\i$}};
    \draw [thin,->,blue] ([yshift=-0.4em]node0\i.center) -- ([yshift=-0.4em,xshift= \resseg]node0\i.center) -- ([yshift=0.3em,xshift= \resseg]node0\i.center) -- ([yshift=0.3em,xshift=0.1em]node0\i.center);
}

\draw [->,very thick] (0+\subseg,0) -- (0+\subseg,11em);
\node [anchor=north] (cap11) at (0+\subseg-0.5em,0) {\footnotesize{2 ODE blocks with}};
\node [anchor=north west] (cap12) at ([yshift=0.3em]cap11.south west) {\footnotesize{3rd-order solutions}};

\node [circlenode,fill=teal!50] (node11) at (0+\subseg,0+1*\seg+0.3*\seg) {};
\node [circlenode,fill=teal!50] (node12) at (0+\subseg,0+2*\seg) {};
\node [circlenode,fill=teal!50] (node13) at (0+\subseg,0+3*\seg-0.3*\seg) {};
\node [circlenode,fill=red] (node14) at (0+\subseg,0+4*\seg+0.3*\seg) {};
\node [circlenode,fill=red] (node15) at (0+\subseg,0+5*\seg) {};
\node [circlenode,fill=red] (node16) at (0+\subseg,0+6*\seg-0.3*\seg) {};

\node [odenode,minimum height=2.6em+\seg] (layer11) at (node12) {};
\node [odenode,minimum height=2.6em+\seg] (layer12) at (node15) {};
\node [anchor=north west] (label11) at ([xshift=0.2em]layer11.north west) {\footnotesize{$\theta_1$}};
\node [anchor=north west] (label12) at ([xshift=0.2em]layer12.north west) {\footnotesize{$\theta_2$}};
\draw [thin,->,blue] ([yshift=-0.6em]node11.center) -- ([yshift=-0.6em,xshift=\resseg]node11.center) -- ([yshift=0.6em,xshift=\resseg]node13.center) -- ([yshift=0.6em,xshift=0.1em]node13.center);
\draw [thin,-,blue] ([yshift=0.3*\seg]node11.center)  -- ([yshift=0.3*\seg,xshift=\resseg]node11.center);
\draw [thin,-,blue] ([yshift=0.3*\seg]node12.center)  -- ([yshift=0.3*\seg,xshift=\resseg]node12.center);
\draw [thin,->,blue] ([yshift=-0.6em]node14.center) -- ([yshift=-0.6em,xshift= \resseg]node14.center) -- ([yshift=0.6em,xshift=\resseg]node16.center) -- ([yshift=0.6em,xshift=0.1em]node16.center);
\draw [thin,-,blue] ([yshift=0.3*\seg]node14.center)  -- ([yshift=0.3*\seg,xshift=\resseg]node14.center);
\draw [thin,-,blue] ([yshift=0.3*\seg]node15.center)  -- ([yshift=0.3*\seg,xshift=\resseg]node15.center);

\node [anchor=west,circlenode,fill=black] (lengend01) at ([xshift=-0.9*\subseg]node06.west) {};
\node [anchor=west] (lengend02) at ([xshift=0em]lengend01.east) {\small{:\ \ $F(\cdot,\theta_t)$}};
\node [anchor=north east,draw,thick,inner sep=0,minimum height=0.8em,minimum width=1em,rounded corners=0.1em] (lengend03) at ([yshift=-1.2em]lengend01.south east) {};
\node [anchor=west] (lengend04) at ([xshift=0em]lengend03.east) {\small{:\ \ ODE}};
\node [anchor=north west] (lengend05) at ([xshift=0em,yshift=0.1em]lengend04.south west) {\small{\ \ \ Block}};

\end{tikzpicture}
\caption{Models with different ODE blocks. }
\label{fig:ode-blocks}
\end{figure}

In this paper we continue the line of research on the ODE-inspired method. The basic idea is to use a high-order method for more accurate numerical solutions to the ODE. This leads to a larger ODE block that generates a sequence of intermediate approximations to the solution. We find that the larger ODE block is sufficient to take the role of several ODE blocks with first-order solutions. The benefit is obvious: the use of fewer ODE blocks lowers the risk of introducing errors in block switching, and the high-order method reduces the approximation error in each ODE block. See Figure \ref{fig:ode-blocks} for a comparison of different models.

Our method is parameter-efficient because $\theta(t)$ is re-used within the same ODE block. As another ``bonus", the model can be improved by learning coefficients of different intermediate approximations in a block. We evaluate our method in strong Transformer systems, covering both the wide (and big) model and the deep model. It achieves 30.76 and 44.11 BLEU scores on the WMT14 En-De and En-Fr test sets. This result sets a new state-of-the-art on the WMT14 En-Fr task.

\section{Transformer and ODEs}

We start with a description of Transformer, followed by its relationship with ODEs. We choose Transformer for our discussion and experiments because it is one of the state-of-the-art models in recent MT evaluations.

\subsection{Transformer}

Transformer is an example of the encoder-decoder paradigm \cite{vaswani2017attention}. The encoder is a stack of identical layers. Each layer consists of a self-attention block and a feedforward network (FFN) block. Both of them equip with a residual connection and a layer normalization unit. Note that the term ``block'' is used in many different ways. In this paper, the term refers to any neural network that is enhanced by the residual connection (occasionally call it a \textit{residual block}).

Following the Pre-norm architecture \cite{wang-etal-2019-learning}, we define a block as

\begin{eqnarray}
y_{t+1} & = & y_t + G(\textrm{LN}(y_t), \theta_t) \label{eq:prenorm}
\end{eqnarray}

\noindent where $\textrm{LN}(\cdot)$ is the layer normalization function\footnote{We drop the parameter of $\textrm{LN}(\cdot)$ for simplicity.}, and $G(\cdot)$ is either the self-attention or feedforward network. The decoder shares a similar architecture, having an additional encoder-decoder attention block sandwiched between the self-attention and FFN blocks.

\subsection{Ordinary Differential Equations}

An ordinary differential equation is an equation involving a function $y(t)$ of a variable $t$ and its derivatives. A simple form of ODE is an equation that defines the first-order derivative of $y(t)$, like this

\begin{eqnarray}
\frac{\mathrm{d} y(t)}{\mathrm{d} t} & = & f(y(t),t) \label{eq:ode-original}
\end{eqnarray}

\noindent where $f(y(t),t)$ defines a time-dependent vector field if we know its value at all points of $y$ and all instants of time $t$. Eq. (\ref{eq:ode-original}) covers a broad range of problems, in that the change of a variable is determined by its current value and a time variable $t$.

This formulation also works with Pre-norm Transformer blocks. For notational simplicity, we re-define $G(\textrm{LN}(y_t), \theta_t)$ as a new function $F(y_t, \theta_t)$:

\begin{eqnarray}
F(y_t, \theta_t) & = & G(\textrm{LN}(y_t), \theta_t)) \label{eq:g-f}
\end{eqnarray}

\noindent We then relax $y_t$ and $\theta_t$ to continuous functions $y(t)$ and $\theta(t)$, and rewrite Eq. (\ref{eq:prenorm}) to be:

\begin{eqnarray}
y(t+\Delta t) & = & y(t) + \Delta t \cdot F(y(t), \theta(t)) \label{eq:prenorm-delta}
\end{eqnarray}

\noindent where $\Delta t$ is the change of $t$, and is general called \textit{step size}. Obviously, we have $\Delta t = 1$ in Transformer. But we can adjust step size $\Delta t$ using a limit, and have

\begin{eqnarray}
\lim_{\Delta t \to 0} \frac{y(t+\Delta t) - y(t)}{\Delta t} & = & F(y(t), \theta(t)) \label{eq:prenorm-limit}
\end{eqnarray}

Given the fact that $\lim_{\Delta t \to 0} \frac{y(t+\Delta t) - y(t)}{\Delta t} = \frac{\mathrm{d} y(t)}{\mathrm{d} t}$, Eq. (\ref{eq:prenorm-limit}) is an instance of Eq. (\ref{eq:ode-original}). The only difference lies in that we introduce $\theta(t)$ into the right-hand side of Eq. (\ref{eq:ode-original}).

Then, we say that a Pre-norm Transformer block describes an ODE. It has been found that Eq. (\ref{eq:prenorm}) shares the same form as the Euler method of solving the ODE described in Eq. (\ref{eq:prenorm-limit}) \cite{haberruthotto2017}. This establishes a relationship between Transformer and ODEs, in that, given $F(\cdot,\cdot)$ and learned parameters $\{\theta_t\}$, the forward pass of a multi-block Transformer is a process of running the Euler method for several steps.

\section{The ODE Transformer}

In numerical methods of ODEs, we want to ensure the precise solutions to the ODEs in a minimum number of computation steps. But the Euler method is not ``precise'' because it is a first-order method, and naturally with local truncation errors. The global error might be larger if we run it for a number of times\footnote{The global error is what we would ordinarily call the error: the difference between $y(t)$ and the true solution. The local error is the error introduced in a single step: the difference between $y(t)$ and the solution obtained by assuming that $y(t-1)$ is the true solution}. This is obviously the case for Transformer, especially when the multi-layer neural network arises a higher risk of unstability in solving the ODEs \cite{haberruthotto2017}.

\subsection{High-Order ODE Solvers}

Here we use the Runge-Kutta methods for a higher order solution to ODEs \cite{runge1895numerische,kutta1901beitrag,butcher1996history,ascher1998computer}. They are a classic family of iterative methods with different orders of precision\footnote{A $p$-order numerical method means that the global truncation error is proportional to $p$ power of the step size.}. More formally, the explicit Runge-Kutta methods of an $n$-step solution is defined to be:

\begin{eqnarray}
y_{t+1} & = & y_{t} + \sum_{i=1}^{n}\gamma_i F_i \label{eq:rk-yt1} \\
    F_1 & = & h f(y_{t},t) \label{eq:rk-y1}\\
    F_i & = & h f(y_{t} + \sum_{j=1}^{i-1} \beta_{ij}F_j, t + \alpha_i h) \label{eq:rk-fi}
\end{eqnarray}

\noindent where $h$ is the step size and could be simply 1 in most cases. $F_i$ is an intermediate approximation to the solution at step $t + \alpha_i h$. $\alpha$, $\beta$ and $\gamma$ are coefficients which can be determined by the Taylor series of $y_{t+1}$ \cite{butcher1963coefficients}. Eq. (\ref{eq:rk-fi}) describes a sequence of solution approximations $\{F_1,...,F_n\}$ over $n$ steps $\{t+\alpha_1 h,...,t+\alpha_n h\}$. These approximations are then interpolated to form the final solution, as in Eq. (\ref{eq:rk-yt1}).

The Runge-Kutta methods are straightforwardly applicable to the design of a Transformer block. All we need is to replace the function $f$ (see Eq. (\ref{eq:rk-fi})) with the function $F$ (see Eq. (\ref{eq:g-f})). The advantage is that the function $F$ is re-used in a block. Also, the model parameter $\theta_t$ can be shared within the block\footnote{Although we could distinguish the parameters at different steps in a block, we found that it did not help and made the model difficult to learn.}. In this way, one can omit $t+\alpha_i h$ in Eq. (\ref{eq:rk-fi}), and compute $F_i$ by

\begin{eqnarray}
F_i & = & F(y_{t} + \sum_{j=1}^{i-1} \beta_{ij}F_j, \theta_t)
\end{eqnarray}

\noindent This makes the system more parameter-efficient. As would be shown in our experiments, the high-order Runge-Kutta methods can learn strong NMT systems with significantly smaller models.

The Runge-Kutta methods are general. For example, the Euler method is a first-order instance of them. For a second-order Runge-Kutta (RK2) block, we have

\begin{eqnarray}
y_{t+1} & = & y_{t} + \frac{1}{2} (F_1 + F_2) \label{eq:rk2} \\
    F_1 & = & F(y_t, \theta_t) \\
    F_2 & = & F(y_t + F_1, \theta_t)
\end{eqnarray}

\noindent This is also known as the improved Euler method. Likewise, we can define a fourth-order Runge-Kutta (RK4) block to be:

\begin{eqnarray}
y_{t+1} & = & y_{t} + \nonumber \\
        &   & \frac{1}{6} (F_1 + 2F_2+ 2F_3+ F_4) \label{eq:rk4} \\
    F_1 & = & F(y_t, \theta_t) \\
    F_2 & = & F(y_t + \frac{1}{2}F_1, \theta_t)\\
    F_3 & = & F(y_t + \frac{1}{2}F_2, \theta_t)\\
    F_4 & = & F(y_t + F_3, \theta_t)
\end{eqnarray}

\definecolor{diy_red}{rgb}{0.996,0.5059,0.4902}
  \tikzstyle{input} = [rectangle, minimum width = 0.6cm, minimum height = 0.3cm, text centered]
  \tikzstyle{output} = [rectangle, minimum width = 0.6cm, minimum height = 0.3cm, text centered]
  \tikzstyle{function} = [rectangle, rounded corners=2pt, minimum width = 0.5cm, minimum height = 0.2cm, text centered, draw = black, thick,fill = diy_red!80]
  \tikzstyle{addition} = [circle, thick, minimum size = 0.2cm, text centered, draw = black]
  \tikzstyle{arrow} = [thick, ->, >=stealth]
  \tikzstyle{line} = [thick, -]
  \tikzstyle{point}=[coordinate]
  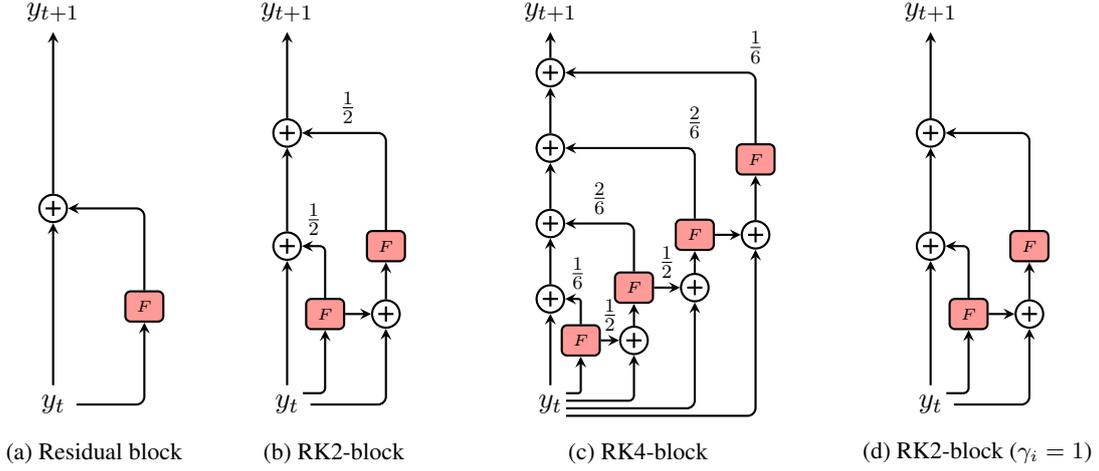
\begin{figure*}
  \centering
  \begin{tikzpicture}[node distance = 0cm]
  \begin{scope}
  \node(output)[output]{$y_{t+1}$};
  \node(addition1)[addition,below of = output,yshift = -2.6cm]{};
  \node(f1)[function,below of = addition1,xshift = 1.2cm,yshift = -1.3cm]{\tiny $F$};
  \node(input)[input,below of = addition1,yshift = -2.6cm]{$y_{t}$};

  \node(point1)[point,below of = addition1,xshift = 0.1cm, yshift = 0cm,align=center]{};
  \node(point2)[point,above of = addition1,xshift = -0.1cm, yshift = 0cm,align=center]{};
  \node(point3)[point,right of = addition1,xshift = 0cm, yshift = 0.1cm,align=center]{};
  \node(point4)[point,left of = addition1,xshift = 0cm, yshift = -0.1cm,align=center]{};

  \draw [arrow] (input) -- (addition1);
  \draw [arrow, rounded corners=3pt] (input.east) -| (f1.south);
  \draw [arrow, rounded corners=3pt] (f1.north) |- (addition1);
  \draw [arrow] (addition1) -- (output);
  \draw [line] (point1) -- (point2);
  \draw [line] (point3) -- (point4);

  \node[font=\footnotesize] at ([xshift=1.4em,yshift=-1em]input.south) { (a) Residual block};
  \end{scope}

  \begin{scope}[xshift=8em]
  \node(output)[output]{$y_{t+1}$};
  \node(addition1)[addition,below of = output,yshift = -1.6cm]{};
  \node(addition2)[addition,below of = addition1,yshift = -1.5cm]{};

  \node(f1)[function,below of = addition2,xshift = 0.5cm,yshift = -0.9cm]{\tiny $F$};
  \node(addition3)[addition,right of = f1,xshift = 0.8cm]{};
  \node(f2)[function,above of = addition3,yshift = 0.9cm]{\tiny $F$};

  \node(input)[input,below of = addition2,yshift = -2.1cm]{$y_{t}$};

  \node(point1)[point,below of = addition1,xshift = 0.1cm, yshift = 0cm,align=center]{};
  \node(point2)[point,above of = addition1,xshift = -0.1cm, yshift = 0cm,align=center]{};
  \node(point3)[point,right of = addition1,xshift = 0cm, yshift = 0.1cm,align=center]{};
  \node(point4)[point,left of = addition1,xshift = 0cm, yshift = -0.1cm,align=center]{};
  \node(point5)[point,below of = addition2,xshift = 0.1cm, yshift = 0cm,align=center]{};
  \node(point6)[point,above of = addition2,xshift = -0.1cm, yshift = 0cm,align=center]{};
  \node(point7)[point,right of = addition2,xshift = 0cm, yshift = 0.1cm,align=center]{};
  \node(point8)[point,left of = addition2,xshift = 0cm, yshift = -0.1cm,align=center]{};
  \node(point9)[point,below of = addition3,xshift = 0.1cm, yshift = 0cm,align=center]{};
  \node(point10)[point,above of = addition3,xshift = -0.1cm, yshift = 0cm,align=center]{};
  \node(point11)[point,right of = addition3,xshift = 0cm, yshift = 0.1cm,align=center]{};
  \node(point12)[point,left of = addition3,xshift = 0cm, yshift = -0.1cm,align=center]{};

  \draw [arrow] (input) -- (addition2);
  \draw [arrow, rounded corners=2pt] ([xshift=-0.1cm, yshift=0.15cm]input.east)  -| (f1.south);
  \draw [arrow, rounded corners=2pt] (f1.north) node[above,xshift=-0.15cm,yshift=0.7cm, scale=0.9]{ $\frac {1}{2}$} |- (addition2);
  \draw [arrow] (addition2) -- (addition1);
  \draw [arrow] (addition1) -- (output);
  \draw [arrow] (f1.east) --(f1.east)-- (addition3);
  \draw [arrow, rounded corners=3pt] (input.east) -| (addition3);
  \draw [arrow] (addition3) -- (f2.south);
  \draw [arrow, rounded corners=3pt] (f2.north) |- node[above, xshift=-0.5cm,scale=0.9]{ $\frac {1}{2}$}(addition1);

  \draw [line] (point1) -- (point2);
  \draw [line] (point3) -- (point4);
  \draw [line] (point5) -- (point6);
  \draw [line] (point7) -- (point8);
  \draw [line] (point9) -- (point10);
  \draw [line] (point11) -- (point12);

  \node[font=\footnotesize] at ([xshift=1.6em,yshift=-1em]input.south) { (b) RK2-block};
  \end{scope}

  \begin{scope}[xshift=17em]
  \node(output)[output]{$y_{t+1}$};
  \node(addition1)[addition,below of = output,yshift = -0.8cm]{};
  \node(addition2)[addition,below of = addition1,yshift = -1cm]{};
  \node(addition3)[addition,below of = addition2,yshift = -1cm]{};
  \node(addition4)[addition,below of = addition3,yshift = -1cm]{};

  \node(f1)[function,below of = addition4,xshift = 0.4cm,yshift = -0.55cm]{\tiny $F$};
  \node(addition5)[addition,right of = f1,xshift = 0.7cm]{};
  \node(f2)[function,above of = addition5,yshift = 0.7cm]{\tiny $F$};
  \node(addition6)[addition,right of = f2,xshift = 0.8cm]{};
  \node(f3)[function,above of = addition6,yshift = 0.7cm]{\tiny $F$};
  \node(addition7)[addition,right of = f3,xshift = 0.8cm]{};
  \node(f4)[function,above of = addition7,yshift = 1cm]{\tiny $F$};

  \node(input)[input,below of = addition4,yshift = -1.4cm]{$y_{t}$};

  \node(p1)[point,left of = addition6, yshift = -1.4cm,align=center]{};
  \node(p2)[point,left of = addition7, yshift = -2.25cm,align=center]{};

  \node(point1)[point,below of = addition1,xshift = 0.1cm, yshift = 0cm,align=center]{};
  \node(point2)[point,above of = addition1,xshift = -0.1cm, yshift = 0cm,align=center]{};
  \node(point3)[point,right of = addition1,xshift = 0cm, yshift = 0.1cm,align=center]{};
  \node(point4)[point,left of = addition1,xshift = 0cm, yshift = -0.1cm,align=center]{};
  \node(point5)[point,below of = addition2,xshift = 0.1cm, yshift = 0cm,align=center]{};
  \node(point6)[point,above of = addition2,xshift = -0.1cm, yshift = 0cm,align=center]{};
  \node(point7)[point,right of = addition2,xshift = 0cm, yshift = 0.1cm,align=center]{};
  \node(point8)[point,left of = addition2,xshift = 0cm, yshift = -0.1cm,align=center]{};
  \node(point9)[point,below of = addition5,xshift = 0.1cm, yshift = 0cm,align=center]{};
  \node(point10)[point,above of = addition5,xshift = -0.1cm, yshift = 0cm,align=center]{};
  \node(point11)[point,right of = addition5,xshift = 0cm, yshift = 0.1cm,align=center]{};
  \node(point12)[point,left of = addition5,xshift = 0cm, yshift = -0.1cm,align=center]{};
  \node(point13)[point,below of = addition3,xshift = 0.1cm, yshift = 0cm,align=center]{};
  \node(point14)[point,above of = addition3,xshift = -0.1cm, yshift = 0cm,align=center]{};
  \node(point15)[point,right of = addition3,xshift = 0cm, yshift = 0.1cm,align=center]{};
  \node(point16)[point,left of = addition3,xshift = 0cm, yshift = -0.1cm,align=center]{};
  \node(point17)[point,below of = addition4,xshift = 0.1cm, yshift = 0cm,align=center]{};
  \node(point18)[point,above of = addition4,xshift = -0.1cm, yshift = 0cm,align=center]{};
  \node(point19)[point,right of = addition4,xshift = 0cm, yshift = 0.1cm,align=center]{};
  \node(point20)[point,left of = addition4,xshift = 0cm, yshift = -0.1cm,align=center]{};
  \node(point21)[point,below of = addition6,xshift = 0.1cm, yshift = 0cm,align=center]{};
  \node(point22)[point,above of = addition6,xshift = -0.1cm, yshift = 0cm,align=center]{};
  \node(point23)[point,right of = addition6,xshift = 0cm, yshift = 0.1cm,align=center]{};
  \node(point24)[point,left of = addition6,xshift = 0cm, yshift = -0.1cm,align=center]{};
  \node(point25)[point,below of = addition7,xshift = 0.1cm, yshift = 0cm,align=center]{};
  \node(point26)[point,above of = addition7,xshift = -0.1cm, yshift = 0cm,align=center]{};
  \node(point27)[point,right of = addition7,xshift = 0cm, yshift = 0.1cm,align=center]{};
  \node(point28)[point,left of = addition7,xshift = 0cm, yshift = -0.1cm,align=center]{};

  \draw [arrow] (input) -- (addition4);
  \draw [arrow, rounded corners=1pt] ([xshift=-0.1cm,yshift=0.15cm]input.east) -| (f1.south);
  \draw [arrow, rounded corners=1pt] (f1.north) |-  node[above,pos=0.6,scale=0.9]{$\frac {1}{6}$}(addition4);
  \draw [arrow] (f1.east) -- node[above,pos=0.5,scale=0.9]{$\frac {1}{2}$}(addition5);
  \draw [arrow] (addition4) -- (addition3);
  \draw [arrow] (addition3) -- (addition2);
  \draw [arrow] (addition2) -- (addition1);
  \draw [arrow] (addition1) -- (output);
  \draw [arrow, rounded corners=2pt] ([xshift=-0.1cm, yshift=0.05cm]input.east) -| (addition5);
  \draw [arrow] (addition5) -- (f2.south);
  \draw [arrow, rounded corners=3pt] (f2.north) |- node[above,pos=0.75,scale=0.9]{$\frac {2}{6}$}(addition3);
  \draw [arrow, rounded corners=3pt] ([xshift=-0.1cm, yshift=-0.05cm]input.east) -| (addition6);
  \draw [arrow, rounded corners=3pt] ([xshift=-0.1cm, yshift=-0.15cm]input.east) -| (addition7);
  \draw [arrow, rounded corners=3pt] (f3.north) |- node[above,pos=0.5,scale=0.9]{$\frac {2}{6}$}(addition2);
  \draw [arrow, rounded corners=3pt] (f4.north) |-  node[above,scale=0.9]{$\frac {1}{6}$}(addition1);
  \draw [arrow] (addition6) -- (f3.south);
  \draw [arrow] (addition7) -- (f4.south);
  \draw [arrow] (f2.east) -- node[above,pos=0.5,scale=0.9]{$\frac {1}{2}$}(addition6);
  \draw [arrow] (f3.east) -- (addition7);

  \draw [line] (point1) -- (point2);
  \draw [line] (point3) -- (point4);
  \draw [line] (point5) -- (point6);
  \draw [line] (point7) -- (point8);
  \draw [line] (point9) -- (point10);
  \draw [line] (point11) -- (point12);
  \draw [line] (point13) -- (point14);
  \draw [line] (point15) -- (point16);
  \draw [line] (point17) -- (point18);
  \draw [line] (point19) -- (point20);
  \draw [line] (point21) -- (point22);
  \draw [line] (point23) -- (point24);
  \draw [line] (point25) -- (point26);
  \draw [line] (point27) -- (point28);

  \node[font=\footnotesize] at ([xshift=3em,yshift=-1em]input.south) { (c) RK4-block};
  \end{scope}

  \begin{scope}[xshift=30em]
  \node(output)[output]{$y_{t+1}$};
  \node(addition1)[addition,below of = output,yshift = -1.6cm]{};
  \node(addition2)[addition,below of = addition1,yshift = -1.5cm]{};

  \node(f1)[function,below of = addition2,xshift = 0.5cm,yshift = -0.9cm]{\tiny $F$};
  \node(addition3)[addition,right of = f1,xshift = 0.8cm]{};
  \node(f2)[function,above of = addition3,yshift = 0.9cm]{\tiny $F$};

  \node(input)[input,below of = addition2,yshift = -2.1cm]{$y_{t}$};

  \node(point1)[point,below of = addition1,xshift = 0.1cm, yshift = 0cm,align=center]{};
  \node(point2)[point,above of = addition1,xshift = -0.1cm, yshift = 0cm,align=center]{};
  \node(point3)[point,right of = addition1,xshift = 0cm, yshift = 0.1cm,align=center]{};
  \node(point4)[point,left of = addition1,xshift = 0cm, yshift = -0.1cm,align=center]{};
  \node(point5)[point,below of = addition2,xshift = 0.1cm, yshift = 0cm,align=center]{};
  \node(point6)[point,above of = addition2,xshift = -0.1cm, yshift = 0cm,align=center]{};
  \node(point7)[point,right of = addition2,xshift = 0cm, yshift = 0.1cm,align=center]{};
  \node(point8)[point,left of = addition2,xshift = 0cm, yshift = -0.1cm,align=center]{};
  \node(point9)[point,below of = addition3,xshift = 0.1cm, yshift = 0cm,align=center]{};
  \node(point10)[point,above of = addition3,xshift = -0.1cm, yshift = 0cm,align=center]{};
  \node(point11)[point,right of = addition3,xshift = 0cm, yshift = 0.1cm,align=center]{};
  \node(point12)[point,left of = addition3,xshift = 0cm, yshift = -0.1cm,align=center]{};

  \draw [arrow] (input) -- (addition2);
  \draw [arrow, rounded corners=2pt] ([xshift=-0.1cm, yshift=0.15cm]input.east) -| (f1.south);
  \draw [arrow,, rounded corners=2pt] (f1.north) |- (addition2);
  \draw [arrow] (addition2) -- (addition1);
  \draw [arrow] (addition1) -- (output);
  \draw [arrow] (f1.east) --(f1.east)-- (addition3);
  \draw [arrow, rounded corners=3pt] (input.east) -| (addition3);
  \draw [arrow] (addition3) -- (f2.south);
  \draw [arrow, rounded corners=3pt] (f2.north) |- (addition1);

  \draw [line] (point1) -- (point2);
  \draw [line] (point3) -- (point4);
  \draw [line] (point5) -- (point6);
  \draw [line] (point7) -- (point8);
  \draw [line] (point9) -- (point10);
  \draw [line] (point11) -- (point12);

  \node[font=\footnotesize] at ([xshift=1.6em,yshift=-1em]input.south) { (d) RK2-block ($\gamma_i=1$)};
  \end{scope}

  \end{tikzpicture}
  \caption{Architectures of ODE Transformer blocks.}
  \label{fig:architecture}
  \end{figure*}

See Figure \ref{fig:architecture} for a comparison of different Runge-Kutta blocks. It should be noted that the method presented here can be interpreted from the perspective of representation refinement \cite{iclrGreff2017}. It provides a way for a function to update the function itself. For example, Universal Transformer refines the representation of the input sequence using the same function and the same parameters in a block-wise manner \cite{Dehghani2019universal}. Here we show that inner block refinements can be modeled with a good theoretical support.

\subsection{Coefficient Learning}
\label{sec:coefficient}

In our preliminary experiments, the RK2 and RK4 methods yielded promising BLEU improvements when the model was shallow. But it was found that the improvement did not persist for deeper models. To figure out why this happened, let us review the Runge-Kutta methods from the angle of training. Take the RK2 method as an example. We rewrite Eq. (\ref{eq:rk2}) by substituting $F_1$ and $F_2$, as follow

\begin{eqnarray}
y_{t+1} & = & y_{t} + \frac{1}{2}F(y_{t},\theta_t) + \nonumber \\
        &   & \frac{1}{2}F(y_{t}+F(y_{t},\theta_t),\theta_t) \label{eq:rk2-full}
\end{eqnarray}

Let $\mathcal{E}$ be the loss of training, $L$ be the number blocks of the model, and $y_{L}$ be the model output. The gradient of $\mathcal{E}$ at $y_t$ is

\begin{eqnarray}
\frac{\partial \mathcal{E}}{\partial y_{t}} & = & \frac{\partial \mathcal{E}}{\partial {y_{L}}} \cdot \frac{1}{2^{L-t}} \cdot \prod_{k=t}^{L-1} (1+g_{k}) \label{eq:rk2-gradient}
\end{eqnarray}


\noindent where
\begin{eqnarray}
{ g_{k}} & = & \hspace{-0.5em} \Big( 1+\frac{\partial F(y_{k},\theta_k)}{\partial y_{k}} \Big) \cdot \nonumber \\
                                        &   & \hspace{-0.5em} \Big( 1+\frac{\partial F(y_{k}+F(y_{k},\theta_k),\theta_k)}{\partial y_{k}+F(y_{k},\theta_k)} \Big) \label{eq:g-gradient}
\end{eqnarray}

\noindent Seen from Eq. (\ref{eq:rk2-gradient}), $\frac{\partial \mathcal{E}}{\partial y_{t}}$ is proportional to the factor $\frac{1}{2^{L-t}}$. This leads to a higher risk of gradient vanishing when $L$ is larger.

The problem somehow attributes to the small coefficients of $F_i$, that is, $\gamma_1 = \gamma_2 = \frac{1}{2}$. A natural idea is to empirically set $\gamma_i = 1$ to eliminate the product factor of less than 1 in gradient computation, although this is not theoretically grounded in standard Runge-Kutta methods. We rewrite Eq. (\ref{eq:rk2-full}) with the new coefficients, as follows

\begin{eqnarray}
y_{t+1} & = & y_{t} + F(y_{t},\theta_t) + \nonumber \\
        &   & F(y_{t}+F(y_{t},\theta_t),\theta_t) \label{eq:rk2-full-new}
\end{eqnarray}

Then, we have the gradient, like this

\begin{eqnarray}
\frac{\partial \mathcal{E}}{\partial y_{t}} & = & \frac{\partial \mathcal{E}}{\partial y_{L}} \cdot \prod_{k=t}^{L-1} g_{k} \label{eq:rk2-gradient-new}
\end{eqnarray}


This model is easy to optimize because $\frac{\partial \mathcal{E}}{\partial_{y_{L}}}$ can be passed to lower-level blocks with no scales. Note that, the methods here are instances of parameter sharing \cite{Dehghani2019universal}. For example, in each ODE block, we use the same function $F$ with the same parameter $\theta_t$ for all intermediate steps. Setting $\gamma_i = 1$ is a further step towards this because $F_i$ is passed to next steps with the same scale. Here we call it implicit parameter sharing.

\begin{table*}[t]
  \small
  \setlength{\tabcolsep}{2.5pt}
  \centering
  \begin{tabular}{llrrrrrrrr}
  \midrule
  \multirow{2}{*}{\textbf{Model}} & \multirow{2}{*}{\textbf{Layers \ }} & \multicolumn{4}{c }{\textbf{WMT En-De}}  & \multicolumn{4}{c}{\textbf{WMT En-Fr}} \\
  \cmidrule(r){3-6} \cmidrule(r){7-10}
  &  & \bf \#Param & \bf Steps & \bf BLEU & \bf SBLEU & \bf \#Param & \bf Steps & \bf BLEU & \bf SBLEU\\
    \midrule
    \citet{vaswani2017attention} - Transformer            &6-6    &213M  &100K   &28.40       &-         &222M  &300K  &41.00       &-        \\
    \citet{ott-EtAl:2018:WMT} - Scaling NMT               &6-6    &210M  &100K   &29.30       &28.6      &222M  &100K  &43.20       &41.4        \\
    \citet{Dehghani2019universal} - Universal Transformer &-      &-     &-      &28.90       &-         &-     &-     &-           &-        \\
    \citet{lu2019understanding} - MacaronNet              &6-6   &-     &-     &30.20      &-          &-     &-     &-           &-        \\
    \citet{fan2019reducing} - LayerDrop                   &12-6   &286M  &100K   &30.20       &-         &-     &-     &-           &-        \\
    \citet{wu-etal-2019-depth} - Depth growing            &8-8    &270M  &800K   &29.92       &-         &-  &-  &43.27       &-        \\
    \citet{wang-etal-2019-learning} - Transformer-DLCL    &30-6    &137M  &50K    &29.30       &28.6     &-     &-     &-           &-        \\
    \citet{zhang-etal-2019-improving} - Depth-wise Scale  &20-20  &560M  &300K  &29.62       &29.0      &108M  &300K  &40.58       &-        \\
    \citet{wei-etal-2020-multiscale} - Multiscale Collaborative &18-6 &512M     &300K     &30.56      &-          &-     &-     &-           &-        \\
    \citet{liu-etal-2020-understanding} - ADMIN           &60-12  &262M  &250K      &30.01       &29.5      &-     &250K     &43.80       &41.8   \\
    \citet{li-etal-2020-shallow} - SDT                    &48-6   &192M  &50K    &30.21       &29.0      &198M  &100K  &43.28       &41.5     \\
    \citet{zhu2020incorporating} - BERT-fused model       &6-6    &-     &-      &30.71       &-         &-     &-     &43.78       &-        \\

    \midrule
    \multicolumn{10}{c}{\bf Base and Deep Models} \\
    \midrule
    Residual-block                                        &6-6    &61M   &50K    &27.89       &26.8      &69M   &100K   &41.05       &39.1     \\
    RK2-block (learnable $\gamma_i$)                      &6-6    &61M   &50K    &28.86       &27.7      &69M   &100K   &42.31       &40.3     \\
    RK4-block                                             &6-6    &61M   &50K    &29.03       &27.9      &69M   &100K   &42.56       &40.6     \\
    Residual-block                                        &24-6   &118M  &50K    &29.43       &28.3      &123M  &100K   &42.67       &40.6     \\
    RK2-block (learnable $\gamma_i$)                      &24-6   &118M  &50K    &\bf 30.29   &\bf 29.2  &123M  &100K   &\bf 43.48    &\bf 41.5 \\
    RK4-block                                             &24-6   &118M  &50K    &29.80       &28.8      &123M  &100K   &43.28       &41.3     \\
    \midrule
    \multicolumn{10}{c}{\bf Wide Models} \\
    \midrule
    Residual-block-Big                                    &6-6    &211M  &100K    &29.21       &28.1      &221M  &100K  &42.89       &40.9     \\
    RK2-block (learnable $\gamma_i$)                      &6-6    &211M  &100K    &30.53       &29.4      &221M  &100K  &43.59       &41.6     \\
    Residual-block-Big                                    &12-6   &286M  &100K    &29.91       &28.9      &297M  &100K  &43.22       &41.2     \\
    RK2-block (learnable $\gamma_i$)                      &12-6   &286M  &100K    &\bf 30.76   &\bf 29.6  &297M  &100K  &\bf 44.11    &\bf 42.2 \\

  \midrule
\end{tabular}
\caption{Comparison with the state-of-the-arts on WMT En-De and WMT En-Fr tasks. We both report the tokenized BLEU and sacrebleu scores for comparison with previous work.}
\label{tab:main-results}
\end{table*}

Another method of scaling $F_i$ is to learn the coefficients automatically on the training data (with the initial value $\gamma_i = 1$). It helps the system learn the way of flowing $F_i$ in a block. Our experiments show that the automatic coefficient learning is necessary for better results (see Section \ref{sec:exp}).



\section{Experiments}
\label{sec:exp}

\subsection{Experimental Setups}
Our proposed methods were evaluated on three widely-used benchmarks: the WMT'14 English-German (En-De), WMT'14 English-French (En-Fr) and WMT'16 English-Romanian (En-Ro) translation tasks.

\paragraph{Datasets and Evaluations:}
For the En-De task, the training data consisted of approximately $4.5$M tokenized sentence pairs, as in \cite{vaswani2017attention}. All sentences were segmented into sequences of sub-word units \cite{sennrich-subword-neural} with $32$K merge operations using a shared vocabulary. We selected \textit{newstest2013} as the validation data and \textit{newstest2014} as the test data. For the En-Fr task, we used the dataset provided by Fairseq, i.e., 36M training sentence pairs from WMT'14. \textit{newstest2012+newstest2013} was the validation data and \textit{newstest2014} was the test data. For the En-Ro task, we replicated the setup of \cite{mehta2020delight}, which used $600$K/$2$K/$2$K sentence pairs for training, evaluation and inference, respectively.


We measured performance in terms of BLEU \cite{papineni-etal-2002-bleu}. Both tokenized BLEU scores \footnote{Computed by \textit{multi-bleu.perl}} and sacrebleu\footnote{BLEU+case.mixed+numrefs.1+smooth.exp+\\tok.13a+version.1.2.12} were reported on the En-De and the En-Fr tasks. Also, we report tokenized BLEU scores on the En-Ro task. The beam size and length penalty were set to $4$ and $0.6$ for the En-De and the En-Fr, and $5$ and $1.3$ for the En-Ro.

\paragraph{Training Details:}
As suggested in \citet{li-etal-2020-shallow}'s work, we used relative positional representation (RPR) for a stronger baseline \cite{shaw-etal-2018-self}. All experiments were trained on $8$ GPUs, with $4,096$ tokens on each GPU. For the En-De and the En-Fr tasks, we employed the gradient accumulation strategy with a step of $2$ and $8$, respectively. We used the Adam optimizer \cite{kingma2014adam} whose hyperparameters were set to $(0.9, 0.997)$, and the max point of the learning rate was set to $0.002$ for fast convergence. We regard merging SAN and FFN as the default ODE block.
More details could be found in our supplementary materials.


\subsection{Results}

\paragraph{Results of En-De and En-Fr:}
Table \ref{tab:main-results} compares ODE Transformer with several state-of-the-art systems. Both RK2-block and RK4-block outperform the baselines by a large margin with different model capacities. For example, RK2-block obtains a $0.97$ BLEU improvement with the base configuration when the depth is $6$. RK4-block yields a gain of +$0.17$ BLEU points on top of RK2-block. This observation empirically validates the conjecture that high-order ODE functions are more efficient. When we switch to deep models, RK2-block is comparable with a 48-layer strong system reported in \cite{li-etal-2020-shallow} with significantly fewer parameters, indicating our method is parameter efficient.

\begin{table}[t]
  \setlength{\tabcolsep}{1.5pt}
  \small
  \centering
  \begin{tabular}{lrrr}
  \toprule
  \bf Model & \bf Params & \bf Epochs & \bf BLEU \\
  \midrule
  Transformer in \citet{mehta2020delight}        & 62M & 170 & 34.30 \\
  DeLight \cite{mehta2020delight}                & 53M & 170 & 34.70 \\
  Int Transformer$^{\dag}$\cite{lin2020towards}  & -   & -   & 32.60 \\
  Transformer (Our impl.)                        & 69M & 20  & 33.49 \\
  RK2-block (learnable $\gamma_i$)               & 69M & 20  & 34.94 \\
  RK2-block-Big (learnable $\gamma_i$)           & 226M& 20  & \textbf{35.28} \\
  \bottomrule
  \end{tabular}
  \caption{Results on the WMT En-Ro task. $\dag$ indicates the related information is not reported.}
  \label{tab:en-ro}
\end{table}

Wide models can also benefit from the enlarging layer depth \cite{wei-etal-2020-multiscale,li-etal-2020-shallow}. The RK-2 ODE Transformer achieves BLEU score of $30.76$ and $44.11$ on the En-De and the En-Fr tasks, significantly surpassing the standard Big model by $1.32$ and $0.70$ BLEU points. This sets a new state-of-the-art on these tasks with fewer parameters. Note that more results on RK4-block (learnable $\gamma_i$) will be reported.

\paragraph{Results of En-Ro:}
Table \ref{tab:en-ro} exhibits model parameters, total training steps and BLEU scores of several strong systems on the En-Ro task. Again, ODE Transformer outperforms these baseline.  As stated in \cite{mehta2020delight}, they trained the model up to $170$ epochs and obtained a BLEU score of $34.70$ through the \texttt{DeLight} model. However, the observation here is quite different. The validation perplexity begins to increase after $20$ epochs. Thus, our baseline is slightly inferior to theirs, but matches the result reported in \citet{lin2020towards}. ODE Transformer achieves even better performance with \texttt{DeLight} within much less training cost. For a bigger model (line 6 in Table \ref{tab:en-ro}), it obtains a BLEU score of $35.28$.

\paragraph{Parameter Efficiency:}
Table \ref{tab:en-de} summaries the results of several efficient Transformer variants, including Lite Transformer \cite{wu2020lite}, DeLight \cite{mehta2020delight} and a light version of the Evolved Transformer \cite{so2019evolved}. As we expected, the proposed ODE Transformer is promising for smaller models. It is comparable in BLEU with \texttt{DeLight} but having $9$M fewer parameters. Under the same model capacity, it outperforms \texttt{DeLight} by $0.84$ BLEU points. These results demonstrate that the proposed method is orthogonal to the model capacity. It may offer a new choice for deploying NMT systems on edge devices.

\subsection{Analysis}
Here we investigate some interesting issues. For simplicity, in the following, we call RK2-block with learnable coefficients as RK2-block-v2.

\begin{figure}[!t]
  \centering
  \begin{tikzpicture}
    \scriptsize{
      \begin{axis}[
	 at={(0,0)},
      ymajorgrids,
      xmajorgrids,
      grid style=dashed,
      width=.28\textwidth,
      height=.24\textwidth,
      legend style={at={(0.61,0.10)}, anchor=south west},
      xlabel={\scriptsize{Encoder Depth}},
      ylabel={\scriptsize{BLEU}},
      ylabel style={yshift=-2em},xlabel style={yshift=0.0em},
      yticklabel style={/pgf/number format/precision=1,/pgf/number format/fixed zerofill},
      ymin=26.5,ymax=30.7, ytick={27.00,28.00,29.00,30.00},
      xmin=4,xmax=40,xtick={6,12,18,24,30,36},
      legend style={yshift=-6pt,xshift=-2em, legend plot pos=right,font={\tiny},cells={anchor=west}}
      ]

      \addplot[blue!60,mark=pentagon*,mark size=1.5pt,thick,mark options={fill=white,draw=blue,line width=0.5pt}] coordinates {(6,27.89) (12,28.68) (18,28.97) (24,29.39) (30,29.52) (36,29.81) };
      \addlegendentry{\scalebox{.8}{Base}}

      \addplot[orange!80,mark=triangle*,,mark size=1.5pt,thick,mark options={fill=white,draw=orange,line width=0.5pt}] coordinates {(6,28.67) (12,29.33) (18,29.58) (24,30.00) (30,29.97) (36,30.11)};
      \addlegendentry{\scalebox{.8}{RK2}}

      \addplot[teal!70,mark=diamond*,mark size=1.5pt,thick,mark options={fill=white,draw=teal,line width=0.5pt}] coordinates {(6,28.89) (12,29.55) (18,29.99) (24,30.20) (30,30.31) (36,30.42) };
      \addlegendentry{\scalebox{.8}{RK2-v2}}

      \addplot[red!60,mark=square*,mark size=1.2pt,thick,mark options={fill=white,draw=red,line width=0.5pt}] coordinates {(6,29.05) (12,29.51) (18,29.76) (24,29.80) (30,29.88) (36,29.89)
      };
      \addlegendentry{\scalebox{.8}{RK4}}
      \end{axis}
     }

	\scriptsize{
      \begin{axis}[
	 at={(16em,0)},
      ymajorgrids,
      xmajorgrids,
      grid style=dashed,
      width=.28\textwidth,
      height=.24\textwidth,
      legend style={at={(0.47,0.13)}, anchor=south west},
      xlabel={\scriptsize{Number of Parameters (M)}},
      ylabel={\scriptsize{BLEU}},
      ylabel style={yshift=-2em},xlabel style={yshift=0.0em},
      yticklabel style={/pgf/number format/precision=1,/pgf/number format/fixed zerofill},
      ymin=24,ymax=29.5, ytick={24.00, 25.00,26.00,27.00,28.00,29.00},
      xmin=10,xmax=130,xtick={20,40,60,80,100,120},
      legend style={yshift=-6pt,xshift=-2em, legend plot pos=right,font={\tiny},cells={anchor=west}}
      ]
      \draw[|-|,line width=0.6pt, black!80, dashed, thick] (62,29.23) -- (110, 29.23);
      \addplot[blue!60,mark=pentagon*,mark size=1.5pt,thick,mark options={fill=white,draw=blue,line width=0.5pt}] coordinates { (19,24.83) (29,26.66) (42,27.73) (62,28.69) (110, 29.03)
      };
      \addlegendentry{\scalebox{.8}{Baseline}}

      \addplot[teal!70,mark=diamond*,mark size=1.5pt,thick,mark options={fill=white,draw=teal,line width=0.5pt}] coordinates {(19, 25.96)(29,27.84) (33,28.24) (42,28.88) (62,29.23) };
      \addlegendentry{\scalebox{.8}{RK2-block-v2}}


      \end{axis}

     }
  \end{tikzpicture}
  \caption{The comparison of BLEU against different encoder depth and the number of model parameters.}\label{fig:encoder-detph}
\end{figure}
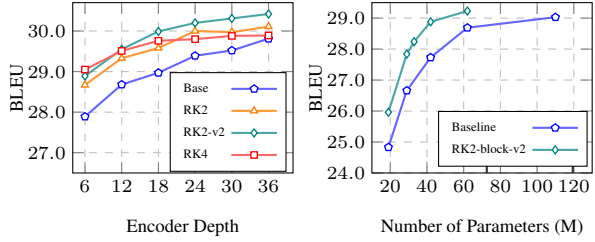

\paragraph{BLEU against Encoder Depth:}
Figure \ref{fig:encoder-detph} (left) depicts BLEU scores of several ODE Transformer variants and the baseline under different encoder depths. All ODE Transformer variants are significantly superior to the baseline when the depth $\leq 24$ . And the RK2-block-v2 almost achieves the best performance over all depths, especially when the model becomes deeper. Intuitively, a 6-layer RK2-block is able to deliver comparable performance compared with the 18-layer baseline system. Again, it indicates the proposed method is parameter efficient. Another finding here is RK4-block behaves strong on shallow models, similar phenomena are observed in Table \ref{tab:main-results}. It is inferior to RK2-block for deeper models, though high-order ODE solvers can obtain lower errors. This is due to original coefficients may cause the optimization problem in the backward propagation when the model is deep (see Section \ref{sec:coefficient}). Also, Figure \ref{fig:encoder-detph} (right) plots BLEU as a function of the model size when the hidden size is $256$. Our RK2 method significantly surpasses the baseline using much fewer parameters.

\begin{table}[t]
  \setlength{\tabcolsep}{1.5pt}
  \small
  \centering
  \begin{tabular}{lrr}
  \toprule
  \bf Model & \bf Params &  \bf BLEU \\
  \midrule
  Transformer \cite{vaswani2017attention}           & 62M  & 27.30 \\
  Evolved Transformer \cite{so2019evolved}          & 46M  & 27.70 \\
  Lite Transformer$^{\dag}$ \cite{wu2020lite}       & -    & 26.50 \\
  DeLight \cite{mehta2020delight}                   & 37M  & 27.60 \\
  RK2-block (learnable $\gamma_i$)                  & 37M  & \textbf{28.24} \\
  RK2-block (learnable $\gamma_i$)                  & 29M  & 27.84 \\
  \bottomrule
  \end{tabular}
\caption{The comparison of model efficiency on the WMT En-De task.}
\label{tab:en-de}
\end{table}

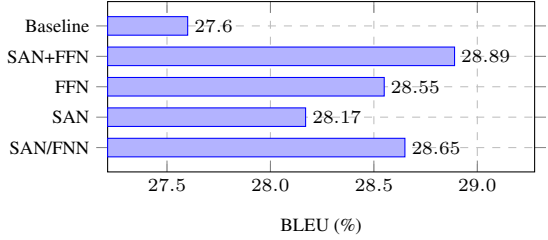
\begin{figure}[!t]
  \centering
  \begin{tikzpicture}
  \scriptsize{
  \begin{axis}[
    ymajorgrids,
    xmajorgrids,
    grid style=dashed,
    xbar,
    height=.24\textwidth,
    width=.45\textwidth,
    bar width=1em,
    xlabel={BLEU (\%)},
    symbolic y coords={{SAN/FNN}, {SAN}, {FFN}, {SAN+FFN}, {Baseline}},
    ytick=data,
    nodes near coords,
    nodes near coords align={horizontal},
    enlarge y limits=0.2,
    enlarge x limits=0.3,xticklabel style={/pgf/number format/fixed,/pgf/number format/fixed zerofill,/pgf/number format/precision=1},]
    \addplot[fill=blue!30, draw=blue] coordinates {(28.65,{SAN/FNN}) (28.17,{SAN}) (28.55,{FFN}) (28.89,{SAN+FFN}) (27.60,{Baseline})};
  \end{axis}
}
  \end{tikzpicture}
  \caption{BLEU scores $[\%]$ of several $F(\cdot,\cdot)$ on the WMT En-De task.}
  \label{fig:components}
\end{figure}

\paragraph{Ablation Study on Different $F(\cdot,\cdot)$:}
As we stated, the $F(\cdot,\cdot)$ function can either be the sublayer, e.g. SAN, FFN or both of them (SAN+FFN). As shown in Figure \ref{fig:components}, high-order ODE works better with FFN than SAN. An exploration might be that the FFN component has more parameters than the SAN component\footnote{Mostly, there are $2 \cdot d_{\mathrm{model}}\cdot 4d_{\mathrm{model}}$ parameters in FFN and $d_{\mathrm{model}}\cdot 3d_{\mathrm{model}}$ + $d_{\mathrm{model}}\cdot d_{\mathrm{model}}$ in SAN.}.
The model that merging FFN and SAN as an ODE block shows the best performance.

\paragraph{Training and Validation Perplexity:}
Figure \ref{fig:loss} plots the training and validation perplexity (PPL) curves of RK blocks and the standard residual-block. We compare the behaviors based on two configurations (base and wide models). Intuitively, RK2-block presents lower training and validation PPLs in both configurations.

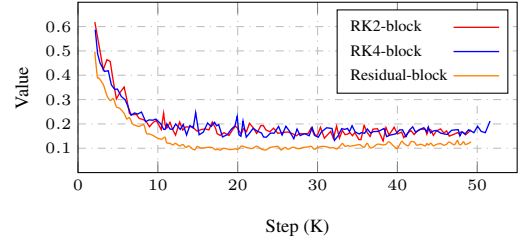
\begin{figure}[!t]
  \centering
  \begin{tikzpicture}
  \scriptsize{
  \begin{axis}[
  width=.46\textwidth, height=.24\textwidth ,
  xlabel=Step (K),
  ylabel=Value,
  xmin=0, xmax=55,
  ymin=0, ymax=0.7,
  xtick={0,10,20,30,40,50},
  ytick={0.1,0.2,0.3,0.4,0.5,0.6},
  yticklabels={$0.1$,$0.2$,$0.3$,$0.4$,$0.5$,$0.6$},
  ymajorgrids=true,
  xmajorgrids=true,
  grid style=dashdotted,
  legend cell align=left,
  scaled ticks=false,
  xlabel style={align=center,font=\scriptsize},
  ylabel style={font=\scriptsize,yshift=-2em},
  y tick style={opacity=0},
  x tick label style={font=\tiny},
  y tick label style={font=\tiny},
  legend style={yshift=-0.2em,xshift=0em,legend cell align=left,legend plot pos=right},
  ]
  \addplot [sharp plot,red,mark size=1pt,thick,line width=0.5pt,mark size=0.2pt] table [x=Step,y=Value,col sep=comma] {Figure/3-2.csv};
  \addplot [sharp plot,blue,mark size=1pt,thick,line width=0.5pt,mark size=0.2pt] table [x=Step,y=Value,col sep=comma] {Figure/3-3.csv};
  \addplot [sharp plot,orange,mark size=1pt,thick,line width=0.5pt,smooth] table [x=Step,y=Value,col sep=comma] {Figure/3-1.csv};
  \legend{\tiny{RK2-block},\tiny{RK4-block},\tiny{Residual-block}},
  \end{axis}
  }
  \end{tikzpicture}
  \caption{Visualization of the gradient norm of ODE Transformers compared with the baseline.}
   \label{fig:gradient_norm}
  \end{figure}

\paragraph{Visualization of the Gradient Norm:}
To study the superiority of the proposed ODE Transformer, we collect the gradient norm of several well-trained systems during training. Figure \ref{fig:gradient_norm} plots the gradient norm of RK2-block, RK4-block and the standard residual-block (baseline). As we can see that Pre-Norm residual block is able to make the training stable \cite{wang-etal-2019-learning}. Both RK2-block and RK4-block provide richer signals due to the implicit parameter sharing among intermediate approximations. And the two learning curves likewise appear to be nearly the same, which is consistent with the result in Table \ref{tab:main-results}.

\paragraph{Comparison of Different ODE Design Schemas:}
Then, we take a comprehensive analysis of several ODE design schemas. As stated in \citet{yiping2018beyond}, several models in computer vision, such as LeapfrogNet \cite{he2019ode}, PolyNet \cite{zhang2017polynet}, Multi-step Net \cite{yiping2018beyond} can also be interpreted from the ODE perspective. The related ODE functions are summarized in Table \ref{tab:comparison}. Here, we re-implement these methods using the same codebase for fair comparisons. We set the encoder depth as $6$ following the base configuration and conducted experiments on the En-De task.


\definecolor{upurple}{RGB}{155,89,182}
\definecolor{ublue}{RGB}{52,152,219}
\definecolor{ured}{RGB}{231,76,60}
\definecolor{udark}{RGB}{77,153,77}
\definecolor{ugreen}{RGB}{46,204,113}
\begin{figure}[!t]
\centering
\begin{tikzpicture}
\scriptsize{
\begin{axis}[
at={(0,0)},
width=.28\textwidth, height=.24\textwidth ,
xtick={0,2,6,...,22},
ytick={4.00,8.00,...,18.00},
xlabel={Epoch},
grid style=dashed,
ylabel={Training\ \ PPL},
xlabel style={align=center,font=\scriptsize},
ylabel style={font=\scriptsize,yshift=-2em},
y tick style={opacity=0},
y tick label style={font=\tiny},
ymajorgrids=true,
xmajorgrids=true,
tick align=inside,
legend pos=outer north east,
yticklabel style={/pgf/number format/precision=1,/pgf/number format/fixed zerofill},
legend style={yshift=-0.5em,xshift=-8.5em,legend cell align=left,legend plot pos=right},
ymin=4.00,
ymax=18.00]

\addplot [sharp plot,orange,smooth,thick,line width=0.5pt,mark=pentagon*,mark size=1pt,thick,mark options={fill=white,draw=orange,line width=0.5pt}] coordinates {(0,0)};
\addplot [sharp plot,ublue,smooth,thick,line width=0.5pt,mark=diamond*,mark size=1pt,thick,mark options={fill=white,draw=ublue,line width=0.5pt}] coordinates {(0,0)};
\addplot [sharp plot,ured,smooth,thick,line width=0.5pt,mark=triangle*,,mark size=1pt,thick,mark options={fill=white,draw=ured,line width=0.5pt}] coordinates {(0,0)};
\addplot [sharp plot,udark,smooth,thick,line width=0.5pt,mark=square*,mark size=0.8pt,thick,mark options={fill=white,draw=udark,line width=0.5pt}] coordinates {(0,0)};

\addplot [sharp plot,orange,smooth,thick,line width=0.5pt] table [x=value,y=result,col sep=comma] {Figure/RK2-Big-train.csv};
\addplot [sharp plot,orange,mark=pentagon*,mark size=1pt,line width=0.5pt,only marks,mark options={fill=white,draw=orange,line width=0.5pt}] table [x=value,y=result,col sep=comma] {Figure/RK2-Big-train-new.csv};

\addplot [sharp plot,ublue,smooth,line width=0.6pt] table [x=value,y=result,col sep=comma] {Figure/RPR-Big-train.csv};
\addplot [sharp plot,ublue,mark=diamond*,mark size=1pt,smooth,line width=0.6pt,only marks,mark options={fill=white,draw=ublue,line width=0.5pt}] table [x=value,y=result,col sep=comma] {Figure/RPR-Big-train-new.csv};

\addplot [sharp plot,ured,smooth,thick,line width=0.6pt] table [x=value,y=result,col sep=comma] {Figure/RK2-Base-train.csv};
\addplot [sharp plot,ured,mark=triangle*,mark size=1pt,smooth,thick,only marks,mark options={fill=white,draw=ured,line width=0.5pt}] table [x=value,y=result,col sep=comma] {Figure/RK2-Base-train-new.csv};

\addplot [sharp plot,udark,smooth,line width=0.6pt] table [x=value,y=result,col sep=comma] {Figure/RPR-Base-train.csv};
\addplot [sharp plot,udark,mark=square*,mark size=0.8pt,smooth,only marks,mark options={fill=white,draw=udark,line width=0.5pt}] table [x=value,y=result,col sep=comma] {Figure/RPR-Base-train-new.csv};

\legend{\tiny{RK2-Big},\tiny{RPR-Big},\tiny{RK2-Base},\tiny{RPR-Base}},
\end{axis}
}
\vspace{6cm}
\scriptsize{
\begin{axis}[
at={(15.5em,0)},
width=.28\textwidth, height=.24\textwidth ,
xtick={0,2,6,...,22},
ytick={4.00,5.00,...,8.00},
xlabel={Epoch},
grid style=dashed,
ylabel={Validation\ \ PPL},
xlabel style={align=center,font=\scriptsize},
ylabel style={font=\scriptsize,yshift=-2.5em},
y tick style={opacity=0},
y tick label style={font=\tiny},
ymajorgrids=true,
xmajorgrids=true,
tick align=inside,
legend pos=outer north east,
legend style={yshift=-0.5em,xshift=-8.5em,legend cell align=left,legend plot pos=right},
yticklabel style={/pgf/number format/precision=1,/pgf/number format/fixed zerofill},
ymin=4.00,
ymax=8.50]
\addplot [sharp plot,orange,smooth,thick,line width=0.5pt,mark=pentagon*,mark size=1pt,thick,mark options={fill=white,draw=orange,line width=0.5pt}] coordinates {(0,0)};
\addplot [sharp plot,ublue,smooth,thick,line width=0.5pt,mark=diamond*,mark size=1pt,thick,mark options={fill=white,draw=ublue,line width=0.5pt}] coordinates {(0,0)};
\addplot [sharp plot,ured,smooth,thick,line width=0.5pt,mark=triangle*,,mark size=1pt,thick,mark options={fill=white,draw=ured,line width=0.5pt}] coordinates {(0,0)};
\addplot [sharp plot,udark,smooth,thick,line width=0.5pt,mark=square*,mark size=0.8pt,thick,mark options={fill=white,draw=udark,line width=0.5pt}] coordinates {(0,0)};

\addplot [sharp plot,orange,smooth,thick,line width=0.5pt] table [x=value,y=result,col sep=comma] {Figure/RK2-Big-valid.csv};
\addplot [sharp plot,orange,mark=pentagon*,mark size=1pt,line width=0.5pt,only marks,mark options={fill=white,draw=orange,line width=0.5pt}] table [x=value,y=result,col sep=comma] {Figure/RK2-Big-valid-new.csv};

\addplot [sharp plot,ublue,smooth,line width=0.6pt] table [x=value,y=result,col sep=comma] {Figure/RPR-Big-valid.csv};
\addplot [sharp plot,ublue,mark=diamond*,mark size=1pt,smooth,line width=0.6pt,only marks,mark options={fill=white,draw=ublue,line width=0.5pt}] table [x=value,y=result,col sep=comma] {Figure/RPR-Big-valid-new.csv};

\addplot [sharp plot,ured,smooth,thick,line width=0.6pt] table [x=value,y=result,col sep=comma] {Figure/RK2-Base-valid.csv};
\addplot [sharp plot,ured,mark=triangle*,mark size=1pt,smooth,thick,only marks,mark options={fill=white,draw=ured,line width=0.5pt}] table [x=value,y=result,col sep=comma] {Figure/RK2-Base-valid-new.csv};

\addplot [sharp plot,udark,smooth,line width=0.6pt] table [x=value,y=result,col sep=comma] {Figure/RPR-Base-valid.csv};
\addplot [sharp plot,udark,mark=square*,mark size=0.8pt,smooth,only marks,mark options={fill=white,draw=udark,line width=0.5pt}] table [x=value,y=result,col sep=comma] {Figure/RPR-Base-valid-new.csv};

\legend{\tiny{RK2-Big},\tiny{RPR-Big},\tiny{RK2-Base},\tiny{RPR-Base}},
\end{axis}
}
\end{tikzpicture}
\caption{The comparison of training and validation PPL on base and wide models.}
\label{fig:loss}
\end{figure}
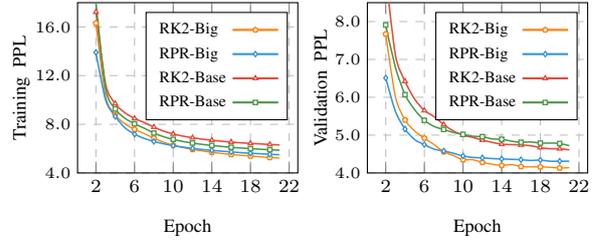

\begin{table*}[t]
  \setlength{\tabcolsep}{2.0pt}
  \renewcommand\arraystretch{1.2}
  \small
  \centering
  \begin{tabular}{llll}
  \toprule
  \bf Model  & \bf Information Flow & \bf Related ODEs &  \bf BLEU \\
  \midrule
  \citet{he2019ode} - Leapfrog                & $y_{t+1} = y_{t-1}+2F(y_{t},\theta_t)$                                                                     & Multistep Euler         & 28.07 \\
  \citet{yiping2018beyond} - Multistep       & $y_{t+1} = k_{n} \cdot y_{t}+(1-k_{n}) \cdot y_{t-1}+F(y_{t},\theta_t)$                                                 & Multistep Euler         & 28.17 \\
  \citet{wang-etal-2019-learning} - DLCL      & $y_{t+1} = y_{0} + \sum_{l=0}^{t}W_{l}{F(y_{l},\theta_l)}$                                                 & Multistep Euler         & 27.78 \\
  \citet{zhang2017polynet} - PolyNet          & $y_{t+1} = y_{t} + F(y_{t},\theta_t) +  F(F(y_{t},\theta_t),\theta_t)$                                     & Backward Euler          & 28.15 \\
  RK2-block                                   & $y_{t+1} = y_{t} + \frac{1}{2}F(y_{t},\theta_t) +  \frac{1}{2}F(y_{t}+F(y_{t},\theta_t),\theta_t)$         & Improved Euler          & 28.57 \\
  RK2-block ($\gamma_i=1$)                    & $y_{t+1} = y_{t} + F(y_{t},\theta_t) +  F(y_{t}+F(y_{t},\theta_t),\theta_t)$                               & Runge-Kutta 2nd-order     & 28.77 \\
  RK2-block (learnable $\gamma_i$)            & $y_{t+1} = y_{t} + \gamma_{1} \cdot F(y_{t},\theta_t)+ \gamma_{2} \cdot F(y_{t}+F(y_{t},\theta_t),\theta_t)$        & Runge-Kutta 2nd-order     & 28.86 \\
  RK4-block                                   & $y_{t+1} = y_{t} + \frac{1}{6}F_{1}+ \frac{2}{6}F_{2} + \frac{2}{6}F_{3} + \frac{1}{6}F_{4}$               & Runge-Kutta 4th-order     & 29.03 \\
  \bottomrule
  \end{tabular}
  \caption{Comparison of several ODE-inspired design schemas on the En-De task. We re-implement and apply these methods into Transformer. Note that $y_{n}$ denotes the model input of layer n. Due to the limited space, we use $F_{i}$ to denote the intermediate representation, where $i \in [1,4]$.}
\label{tab:comparison}
\end{table*}
At time $t$, Multistep Euler methods requires previous states, e.g. $y_{t-1}$, to generate the current approximation, instead of iterative refinements based on the current-time state. Basically, these methods are not parameter efficient, and obtain inferior performance than ours. Note that DLCL can also be regarded as a multistep Euler method, which is more competitive in deep Transformer. But there is only a small improvement upon a shallow baseline.
Theoretically, the Backward Euler method is slightly better than the Forward Euler method in numerical analysis, but the improvement is marginal. Note that our ODE Transformer achieves consistent BLEU improvements over the aforementioned methods. The reason here is that this kind of iterative refinements enable the parameters learning more efficient and effective. All models could be found in our attachment.

\begin{table}[t]
  \setlength{\tabcolsep}{1.5pt}
  \small
  \centering
  \begin{tabular}{lcc}
  \toprule
  \bf Model  & \bf 1-layer PPL & \bf 2-layer PPL \\
  \midrule
  Residual Block                   & 142.33 & 136.07 \\
  RK2-block                        & 131.80 & 123.12 \\
  RK2-block ($\gamma_i=1$)         & 132.67 & 123.90 \\
  RK2-block (learnable $\gamma_i$) & 128.48 & 121.02 \\
  RK4-block                        & \bf 126.89 & \bf 119.46 \\

  \bottomrule
  \end{tabular}
  \caption{The comparison of PPL on several systems. More details refer to the supplementary materials.}
\label{tab:truncate_error}
\end{table}

\paragraph{Quantization of the Truncation Error:}
Here, we aim at quantifying the truncation error. However, we cannot obtain the ``true" solution of each block output in NMT, because we mainly experimented on the encoder side. Instead, we experimented on the language modeling task, where the loss between a single layer model output and the ground truth is equivalent to the truncation error without error propagations.
Table \ref{tab:truncate_error} shows the PPL on the PTB task. All ODE Transformer variants reduce the errors significantly. RK4-order achieves the lowest PPL on both settings. In addition, a RK2-block can even obtain lower PPL than a 2-layer residual-block. The observation here again verifies our conjecture.


\section{Related Work}

\paragraph{Deep Transformer models:}
Recently, deep Transformer has witnessed tremendous success in machine translation. A straightforward way is to shorten the path from upper-level layers to lower-level layers thus to alleviate the gradient vanishing or exploding problems \cite{bapna-etal-2018-training,wang-etal-2019-learning,wu-etal-2019-depth,wei-etal-2020-multiscale}. For deeper models, the training cost is nonnegligible. To speed up the training, an alternative way is to train a shallow model first and progressively increasing the model depth \cite{li-etal-2020-shallow,dong2020towards}.

Apart from the model architecture improvements, another way of easing the optimization is to utilize carefully designed parameter initialization strategies, such as depth-scale \cite{zhang-etal-2019-improving}, Lipschitz constraint \cite{xu-etal-2020-lipschitz}, T-fixup \cite{Huang2020improving} and ADMIN \cite{liu-etal-2020-understanding}.
Note that the ODE Transformer is orthogonal to the aforementioned methods, and we will test it on these methods in the future work.

\paragraph{Ordinary Differential Equations:}
The relationship between the ResNet and ODEs was first proposed by \citet{weinan2017proposal}. This brings the community a brand-new perspective on the design of effective deep architectures. Some insightful architectures \cite{zhang2017polynet,larsson2017fractalnet,yiping2018beyond,he2019ode} can also be interpreted from the ODE perspective. But, in nature language processing, it is still rare to see studies on designing models from the ODE perspective. Perhaps the most relevant work with us is \citet{lu2019understanding}'s work. They interpreted the Transformer architecture from a multi-particle dynamic system view and relocated the self-attention sandwiched into the FFN. Unlike their work, we argue that the stacked first-order ODE blocks may cause error accumulation, thus hindering the model performance. We address this issue by introducing high-order blocks, and demonstrate significant BLEU improvements.

\section{Conclusions}

In this paper, we have explored the relationship between Transformer and ODEs. We have proposed a new architecture (ODE Transformer) to help the model benefit from high-order ODE solutions. Experimental results show that ODE Transformer can significantly outperform the baseline with the same model capacity. It achieves $30.76$ and $44.11$ BLEU scores on the WMT'14 En-De and En-Fr test data. This sets a new state-of-the-art on the En-Fr task.

  \bibliography{acl2021}
  \bibliographystyle{acl_natbib}

  \appendix

  \section{Details for Experimental Steps and Datasets}

All models were trained on 8 NVIDIA TITAN V GPUs with mix-precision accelerating. And main results were the average of three times running with different random seeds. Note that we averaged the last 5/10 checkpoints for more robust results.

Since the proposed method is orthogonal to the model capacity, we evaluated the ODE Transformer on Base/Deep/Wide configurations, respectively. The detail of each configuration is as follows:

\begin{itemize}

\item Base/Deep Model. The hidden size of self-attention was $512$, and the dimension of the inner-layer in FFN was $2,048$. We used 8 heads for attention. For training, we set all dropout to $0.1$, including residual dropout, attention dropout, ReLU dropout. Label smoothing $\epsilon_{ls}=0.1$ was applied to enhance the generation ability of the model. For deep models, we only enlarged the encoder depth considering the inference speed.

\item Wide (or Big) Model. We used the same architecture as Transformer-Base but with a larger hidden layer size $1,024$, more attention heads ($16$), and a larger feed forward inner-layer ($4,096$ dimensions). The residual dropout was set to $0.3$ for the En-De task and $0.1$ for the En-Fr tasks.
\end{itemize}

In the training phase, Deep/Big models were updated for 50K and 100K steps on the En-De task, 100K steps on the En-Fr task, 17K steps on the En-Ro task.

Table \ref{tab:data} summarizes the details of our datasets, including the WMT En-De, the WMT En-Fr and the WMT En-Ro tasks. We both present the sentences and tokens of each task. For En-De and En-Fr task, the datasets used in this work could be found in \texttt{Fairseq}\footnote{\url{https://github.com/pytorch/fairseq/tree/master/examples/scaling_nmt}}. For En-Ro, one can use the preprocessed dataset provided by \texttt{DeLight}\footnote{\url{https://github.com/sacmehta/delight/blob/master/readme_files/nmt/wmt16_en2ro.md}}. Note that we only share the target embedding and the softmax embedding instead of a shared vocabulary between the source side and the target side.

\begin{table}[t]
  \setlength{\tabcolsep}{8.5pt}
  \small
  \centering
  \begin{tabular}{lrrr}
  \toprule
  \bf Lang  & \bf Train & \bf Valid & \bf Test \\
  \midrule
  WMT'14 En-De         & 4.5M  & 3000    & 3003\\
  WMT'14 En-Fr         & 35.7M & 26822    & 3003\\
  WMT'16 En-Ro         & 602K  & 1999    & 1999\\
  
  \bottomrule
  \end{tabular}
  \caption{The comparison of PPL on several systems.}
\label{tab:data}
\end{table}

\section{Details for the PTB dataset}
Here, we introduce the details about the PTB dataset and the corresponding configuration. It contains 88K, 3370 and 3761 sentences for training, validation and test. The vocabulary size was 10K. In this work, the layer depth of the language model was set to $1$ or $2$. The main concern here is to evaluate the truncate error. Assume the layer depth is $1$, then the loss between the block output and the ground-truth can be regarded as the truncate error. It alleviates the influence of the error accumulation among different layers.

The hidden size was $512$, and the filter size of the FFN was $2,048$. We set all the dropout rate as $0.1$, including the residual dropout, attention dropout and the relu dropout. Each model was trained up to 20 epochs, and most models achieved the lowest PPL on the validation set when the epoch is 10. Then the validation PPL began to increase, though the training PPL is still declining. The warmup-step was 2000 and the batch size was $4,096$. The max learning rate was set to $0.0007$. After warmup, the learning rate decayed proportionally to the inverse square root of the current step.

\section{Derivations of the Equation}

Let $\mathcal{E}$ be the loss of training, $L$ be the number blocks of the model, and $y_{L}$ be the model output. 
Here, we define

\begin{eqnarray}
  z_k  &=& y_k + F(y_k,\theta_k)
\end{eqnarray}

Then the information flow of the RK2 method can be described as follows:

\begin{eqnarray}
  y_{k+1} &=& y_k + \frac{1}{2}F(y_k,\theta_k) + \nonumber \\
  &    &\frac{1}{2}F(y_k+F(y_k,\theta_k),\theta_k)\nonumber \\
  & =  & y_k + \frac{1}{2}F(y_k,\theta_k) + \frac{1}{2}F(z_k,\theta_k) \label{eq:RK2}
\end{eqnarray}

\noindent where $\frac{\partial z_{k}}{\partial y_{k}}= 1 + \frac{\partial F(y_k,\theta_k)}{\partial y_k}$.
In this way, the detail derivation of Eq. (\ref{eq:RK2}) is as follows: 

\begin{eqnarray}
  {\small \hspace{-1em} \frac{\partial y_{k+1}}{\partial y_{k}}} 
  & =  & 1 + \frac{1}{2}\frac{\partial F(y_k,\theta_k)}{\partial y_k} + \frac{1}{2}\frac{\partial F(z_k,\theta_k)}{\partial z_k} \cdot \frac{\partial z_k} {\partial y_k} \nonumber \\
  &  =  &\frac{1}{2} \cdot \Big(1 + 1 +\frac{\partial F(y_k,\theta_k)}{\partial y_k}+\frac{\partial F(z_k,\theta_k)}{\partial z_k} \cdot \nonumber \\
  &    & \Big(1+\frac{\partial F(y_k,\theta_k)}{\partial y_k}\Big)\Big)\nonumber \\
  & =  & \frac{1}{2} \cdot \Big(1 + \Big(1+\frac{\partial F(z_k,\theta_k)}{\partial z_k}\Big) \cdot \nonumber \\
  &    & \Big(1+\frac{\partial F(y_k,\theta_k)}{\partial y_k}\Big)\Big)\label{eq:gk}
  \end{eqnarray}

With the chain rule, the error $\mathcal{E}$ propagates from the top layer $y_L$ to layer $y_t$ by the following formula:
  \begin{eqnarray}
        \frac{\partial \mathcal{E}}{\partial y_t} = \frac{\partial \mathcal{E}}{\partial y_L} \cdot \frac{\partial y_{L}}{\partial y_{L-1}} \cdot \frac{\partial y_{L-1}}{\partial y_{L-2}} \cdots \frac{\partial y_{t+1}}{\partial y_{t}} \nonumber \label{eq:chain}
  \end{eqnarray}

Here we have

\begin{eqnarray}
  {\small \hspace{-1em} g_{k}} & = & \hspace{-0.7em} \Big( 1+\frac{\partial F(y_{k},\theta_k)}{\partial y_{k}} \Big) \cdot \Big(1+\frac{\partial F(z_k,\theta_k)}{\partial z_k} \Big) \label{eq:g-gradient}
  \end{eqnarray}

Then, put the Eq. (\ref{eq:chain}) into Eq. (\ref{eq:gk}), the gradient of $\mathcal{E}$ at $y_t$ is

  \begin{eqnarray}
  \frac{\partial \mathcal{E}}{\partial y_{t}} & = & \frac{\partial \mathcal{E}}{\partial_{y_{L}}} \cdot \frac{1}{2^{L-t}} \cdot \prod_{k=t}^{L-1} (1+g_{k}) \label{eq:rk2-gradient}
  \end{eqnarray}
  
 Similarly, we can easily obtain the gradient of RK2 method where $\gamma_i=1$:

  \begin{eqnarray}
        \frac{\partial \mathcal{E}}{\partial y_t} &=& \frac{\partial \mathcal{E}}{\partial y_L} \cdot g_{L-1} \cdot g_{L-2} \cdots g_{t} \nonumber \\
    &=& \frac{\partial \mathcal{E}}{\partial y_L} \cdot \prod_{k=t}^{L-1}g_k 
  \end{eqnarray}

\end{document}